\documentclass{article}

\usepackage{microtype}
\usepackage{graphicx}
\usepackage{subcaption}
\usepackage{booktabs}
\usepackage{hyperref}

\usepackage[preprint]{icml2026}
\usepackage{amsmath}
\usepackage{amssymb}
\usepackage{mathtools}
\usepackage{amsthm}
\usepackage{algorithm}
\usepackage{algorithmic}
\usepackage{xcolor}
\usepackage[capitalize,noabbrev]{cleveref}

\theoremstyle{plain}

\theoremstyle{definition}

\icmltitlerunning{Open-TQ-Metal: Fused Compressed-Domain Attention on Apple Silicon}

\begin{document}

\twocolumn[
  \icmltitle{Open-TQ-Metal: Fused Compressed-Domain Attention \\ for Long-Context LLM Inference on Apple Silicon}

  \vskip 0.3in

  \begin{center}
    {\large\bfseries Sai Vegasena}

    \vskip 0.15in

    {\normalsize Ensue}

    \vskip 0.05in

    {\small\texttt{sai@ensue.dev}}
  \end{center}

  \icmlcorrespondingauthor{Sai Vegasena}{sai@ensue.dev}

  \vskip 0.3in
]

\printAffiliationsAndNotice{}

\begin{figure*}[t]
\centering
\begin{subfigure}[t]{0.44\textwidth}
    \centering
    \includegraphics[width=\textwidth]{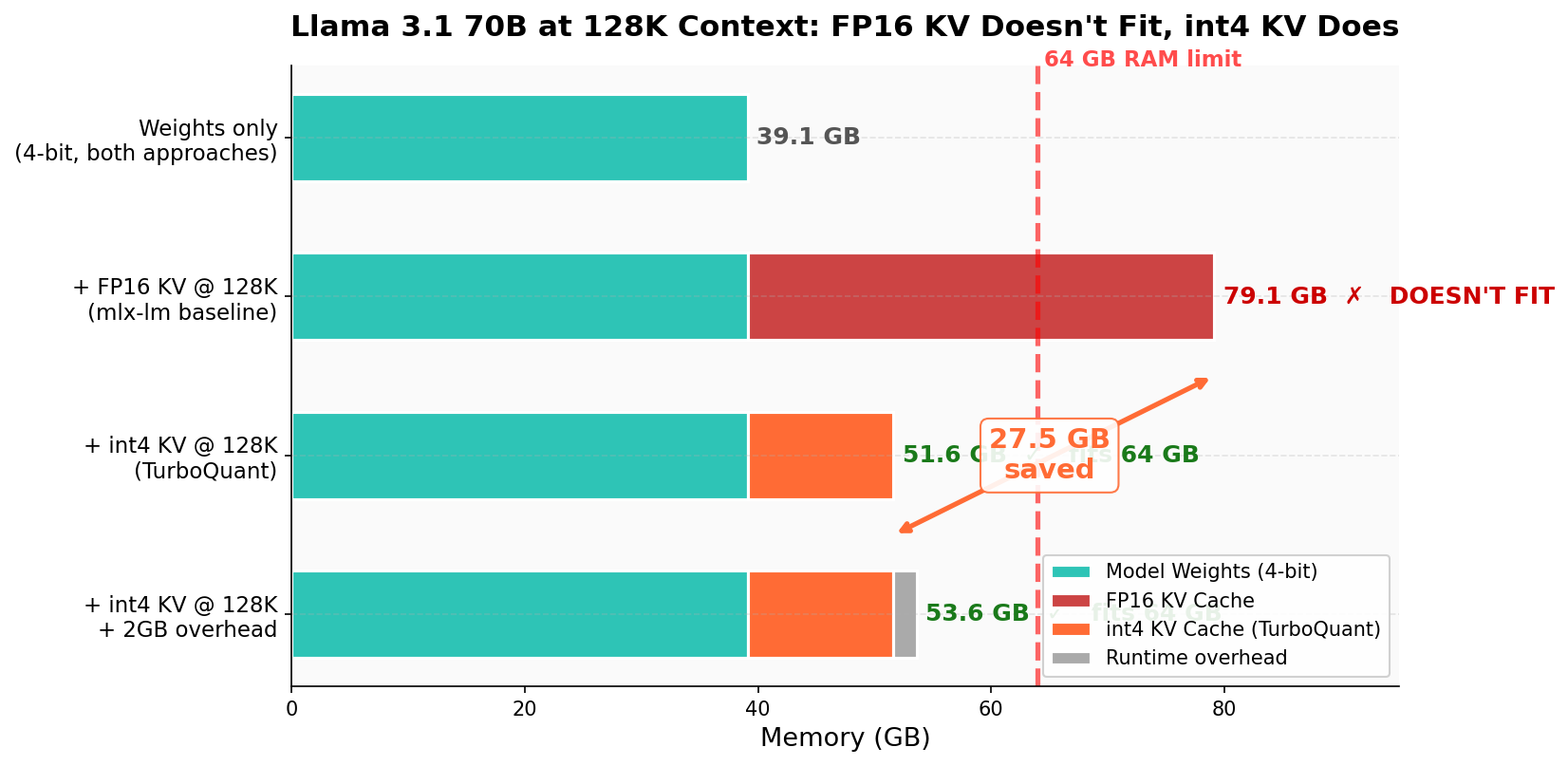}
    \vspace{2pt}
    \caption{Llama~3.1~70B at 128K context: FP16 KV requires 79.1\,GB (infeasible on 64\,GB). Int4 KV reduces total to 53.6\,GB.}
    \label{fig:memory_128k}
\end{subfigure}
\hspace{0.06\textwidth}
\begin{subfigure}[t]{0.44\textwidth}
    \centering
    \includegraphics[width=\textwidth]{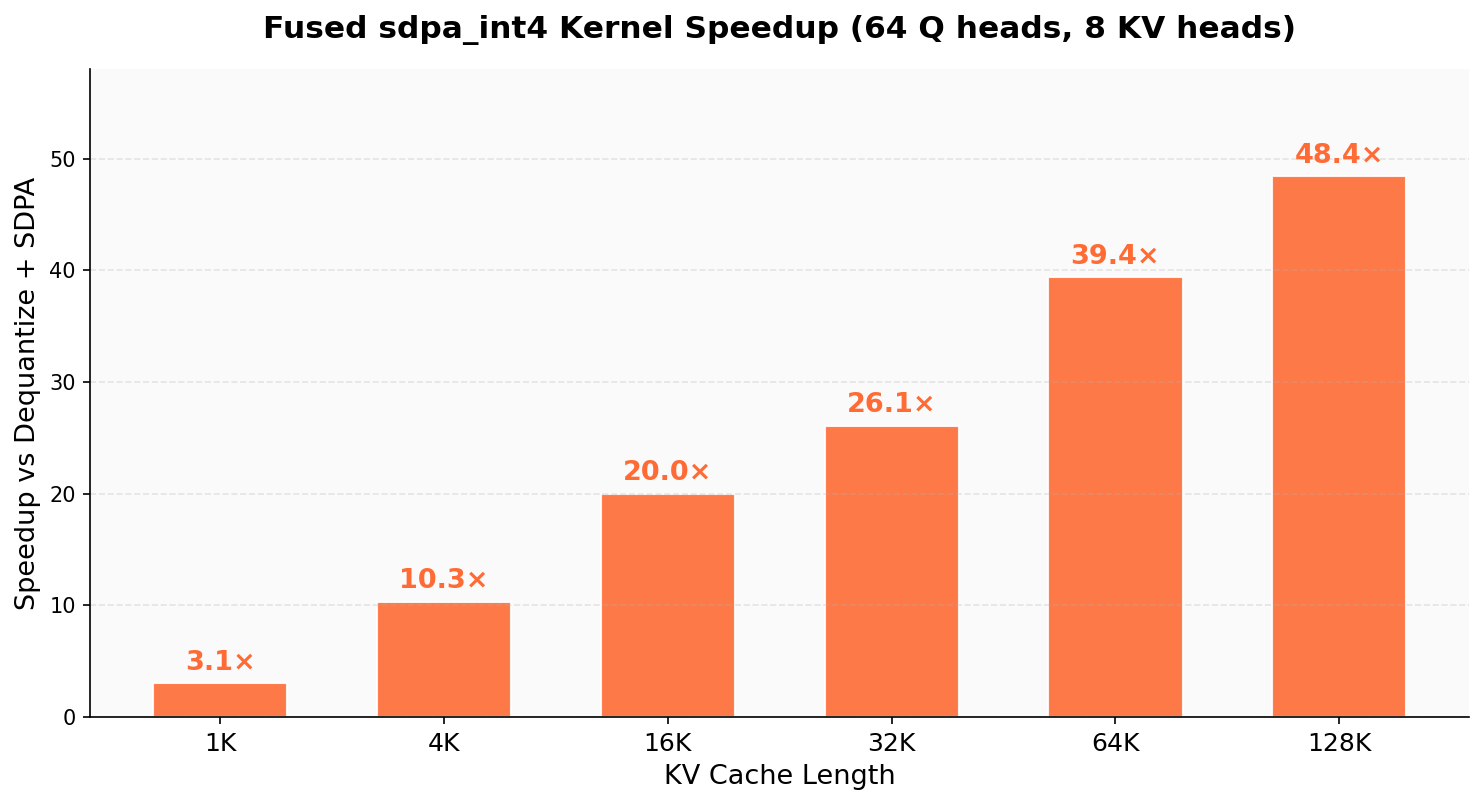}
    \vspace{2pt}
    \caption{Fused \texttt{sdpa\_int4} kernel speedup on Llama~70B. Scales super-linearly from 3$\times$ at 1K to 48$\times$ at 128K.}
    \label{fig:kernel_speedup}
\end{subfigure}
\vspace{4pt}
\caption{The two core results of Open-TQ-Metal: (a)~enabling 128K context on hardware where it was previously impossible, and (b)~achieving super-linear attention speedup via fused compressed-domain computation.}
\label{fig:hero}
\end{figure*}

\begin{abstract}
We present Open-TQ-Metal, the first implementation of fused compressed-domain attention on Apple Silicon, enabling 128K-context inference for Llama~3.1~70B on a single 64GB consumer Mac---a configuration impossible with all existing inference frameworks.
Running large language models at their full designed context window on consumer hardware is blocked by KV cache memory: Llama~3.1~70B at 128K context requires 79.1\,GB in FP16, exceeding the 64\,GB physical limit.
Open-TQ-Metal quantizes the KV cache to int4 on the fly and computes attention directly on the compressed representation via custom Metal compute shaders, eliminating all intermediate dequantization matrices.
Across 330 experiments spanning two model families (Gemma~4~31B and Llama~3.1~70B), the fused \texttt{sdpa\_int4} kernel achieves 48$\times$ attention speedup at 128K context over the dequantize-then-attend baseline, reduces KV cache memory from 40\,GB to 12.5\,GB (3.2$\times$ compression), and maintains identical top-1 token predictions to FP16 inference.
We further provide the first cross-architecture analysis of KV cache quantization methods, revealing that the attention scale factor---not model size---determines whether angular quantization schemes like PolarQuant succeed or fail, with Gemma~4's \texttt{attn\_scale=1.0} amplifying directional error 25--100$\times$ more than Llama's standard $1/\sqrt{d}$ scaling.
\end{abstract}

\section{Introduction}
\label{sec:intro}

Large language models achieve their strongest performance at long context windows~\citep{grattafiori2024llama3}, yet deploying these models on consumer hardware is fundamentally constrained by KV cache memory~\citep{kwon2023vllm}. The KV cache grows linearly with sequence length and model depth: for a transformer~\citep{vaswani2017attention} with $L$ layers and $H_{\text{kv}}$ key-value heads~\citep{shazeer2019mqa,ainslie2023gqa}, Llama~3.1~70B~\citep{grattafiori2024llama3} at 4-bit weights~\citep{frantar2023gptq} occupies 39.1\,GB; its FP16 KV cache at 128K tokens adds 40\,GB more---totaling 79.1\,GB, which exceeds the 64\,GB unified memory of Apple's highest-capacity consumer chips. Every existing inference framework (mlx-lm~\citep{mlx2024}, llama.cpp~\citep{llamacpp}) hits this wall.

Recent KV cache quantization methods~\citep{liu2024kivi,hooper2024kvquant,kang2024gear} compress the cache to 2--4 bits but still dequantize before attention, incurring the full bandwidth cost of materialized temporary matrices. TurboQuant~\citep{turboquant2026} combines MSE-optimal quantization with QJL residual compression, while PolarQuant~\citep{polarquant2026} and QJL~\citep{qjl2025} offer complementary approaches to KV cache compression. However, all three methods have only been evaluated on models up to 8B parameters (32 layers), and none have been validated on Apple Silicon's Metal compute architecture. Their behavior at 70B (80 layers) and across architectures with non-standard attention scaling~\citep{gemma2team2024} is unknown.

We address both gaps with Open-TQ-Metal, a complete inference system that implements fused compressed-domain attention in Metal compute shaders, building on the online softmax technique~\citep{milakov2018online} and split-K parallelism from FlashDecoding~\citep{flashdecoding2023}. Our contributions:

\begin{itemize}
    \item \textbf{Fused int4 SDPA kernel for Metal.} A Metal compute shader that reads packed int4 keys and values directly from device memory, dequantizes per-element in GPU registers via bitwise operations, and computes attention with online softmax---producing zero intermediate matrices. At 128K context, this kernel is 48$\times$ faster than the dequantize-then-attend baseline (\cref{sec:kernel}).

    \item \textbf{128K context on 64\,GB hardware.} By compressing the KV cache from 40\,GB (FP16) to 12.5\,GB (int4), Open-TQ-Metal fits Llama~3.1~70B at 128K context in 53.6\,GB (including 2\,GB runtime overhead)---with 10.4\,GB headroom on a 64\,GB Mac. Output quality is identical: both paths produce the same top-1 token under greedy decode (\cref{sec:results}).

    \item \textbf{Cross-architecture quantization analysis.} Across 330 experiments on Gemma~4~31B and Llama~3.1~70B, we demonstrate that the attention scale factor is the critical variable for quantization method selection: PolarQuant fails on Gemma~4 (\texttt{attn\_scale=1.0}) but succeeds on Llama ($1/\sqrt{d} = 0.0884$), with KL divergence differing by 25--100$\times$ (\cref{sec:analysis}).

    \item \textbf{Split-K parallelism via chained MLX Primitives.} We solve the Metal dispatch race condition for long-context attention by implementing partial and reduce phases as separate MLX Primitives chained through the computation graph, achieving flat latency scaling from 1K to 128K tokens (\cref{sec:splitk}).
\end{itemize}

All code, Metal shaders, and benchmarks are open-sourced. \Cref{fig:hero} summarizes the two core results.

\section{Background}
\label{sec:background}

\subsection{KV Cache Compression}
\label{sec:bg_kv}

Autoregressive inference stores key and value projections from all previous tokens in the \emph{KV cache}~\citep{vaswani2017attention}. For a model with $L$ layers, $H_{\text{kv}}$ KV heads~\citep{shazeer2019mqa,ainslie2023gqa}, head dimension $d$, and sequence length $S$, the cache occupies $2 L H_{\text{kv}} S d b$ bytes, where $b$ is bytes per element. At FP16 ($b=2$), Llama~3.1~70B at 128K context requires:
\begin{equation}
2 \times 80 \times 8 \times 128\text{K} \times 128 \times 2 \;=\; 40\,\text{GB}
\label{eq:kv_size}
\end{equation}
This single allocation exceeds the total memory of most consumer devices.

\textbf{Per-group asymmetric int4 quantization} partitions each vector into groups of $g$ consecutive elements and computes per-group scale and zero-point parameters:
\begin{align}
s &= \frac{\max - \min}{15}, \qquad z = \frac{-\min}{s} \\
q &= \text{clamp}\!\left(\left\lfloor \frac{x}{s} + z \right\rceil, 0, 15\right), \qquad \hat{x} = (q - z) \cdot s
\end{align}
Values are packed two per byte (4 bits each). At $g=32$, the effective compression is 3.2$\times$ including amortized scale and zero-point overhead.

\textbf{PolarQuant}~\citep{polarquant2026} transforms KV embeddings into polar coordinates via a recursive algorithm and quantizes the resulting angles. After random preconditioning, these angles exhibit a tightly concentrated distribution that eliminates the need for explicit normalization. The authors report over 4.2$\times$ compression. However, quantizing angles introduces structured angular error---a property we analyze in \cref{sec:analysis}.

\textbf{QJL}~\citep{qjl2025} projects keys through a random Gaussian matrix and stores only the sign bits of the result, yielding binary sketches in $\{-1,+1\}^m$. Attention scores are estimated asymmetrically: the query side remains in full precision while only keys are binarized. Compression exceeds 5$\times$.

\subsection{TurboQuant and Metal Compute}
\label{sec:bg_tq}

TurboQuant~\citep{turboquant2026} proposes an MSE-optimal scalar quantizer based on random rotation and Lloyd-Max codebooks, with a 1-bit QJL transform applied to the residual. The paper reports quality neutrality at 3.5 bits per dimension and marginal degradation at 2.5 bits, evaluated on models up to 8B parameters.

Apple Silicon GPUs use 32-wide SIMD groups as the basic execution unit. The MLX framework~\citep{mlx2024} provides a C++ Primitive system for dispatching custom Metal kernels. Critically, multiple dispatches within a single \texttt{eval\_gpu()} call execute without guaranteed ordering---a constraint that shapes our split-K design (\cref{sec:splitk}).

\section{Method}
\label{sec:method}

\subsection{Fused int4 Scaled Dot-Product Attention}
\label{sec:kernel}

Standard int4 attention dequantizes the full key matrix to FP32 before computing $\mathbf{Q}\mathbf{K}^T$, materializing a temporary $S \times d$ matrix per head---742\,MB across 50 layers at 950 tokens for Gemma~4. Our fused kernel (\cref{alg:fused}) eliminates this entirely, processing the full KV sequence in a single pass:

\begin{algorithm}[t]
\caption{Fused int4 SDPA (single query, decode step)}
\label{alg:fused}
\begin{algorithmic}[1]
\REQUIRE $\mathbf{q} \in \mathbb{R}^d$; packed int4 $\mathbf{K}, \mathbf{V}$ with scales $\mathbf{s}$, zeros $\mathbf{z}$
\STATE $\mathbf{q} \leftarrow \mathbf{q} / \sqrt{d}$ \hfill $\triangleright$ \textit{pre-scale query}
\STATE $m \leftarrow -\infty$, \; $\ell \leftarrow 0$, \; $\mathbf{o} \leftarrow \mathbf{0}$
\FOR{$i = 1$ \textbf{to} $S$}
    \STATE $\hat{\mathbf{k}}_i \leftarrow \mathrm{dequant4}(\mathbf{K}_i, \mathbf{s}_K, \mathbf{z}_K)$ \hfill $\triangleright$ \textit{in register}
    \STATE $a_i \leftarrow \mathbf{q}^\top \hat{\mathbf{k}}_i$ \hfill $\triangleright$ \textit{SIMD reduction}
    \STATE $m' \leftarrow \max(m,\; a_i)$
    \STATE $\ell \leftarrow \ell \cdot e^{m - m'} + e^{a_i - m'}$
    \STATE $\hat{\mathbf{v}}_i \leftarrow \mathrm{dequant4}(\mathbf{V}_i, \mathbf{s}_V, \mathbf{z}_V)$
    \STATE $\mathbf{o} \leftarrow \mathbf{o} \cdot e^{m - m'} + e^{a_i - m'} \cdot \hat{\mathbf{v}}_i$
    \STATE $m \leftarrow m'$
\ENDFOR
\STATE \textbf{return} $\mathbf{o}\, / \,\ell$
\end{algorithmic}
\end{algorithm}

\noindent where $\mathrm{dequant4}$ extracts 4-bit nibbles via bitwise masks and applies per-group affine reconstruction $(q - z) \cdot s$ entirely in GPU registers---no intermediate matrix is written to device memory.

\textbf{Vectorized nibble extraction.} Rather than extracting nibbles one at a time (4 shifts and masks per uint32), we adopt MLX's \texttt{qdot} pattern: reinterpret the packed uint32 array as uint16, pre-divide query values by $\{1, 16, 256, 4096\}$, and multiply against masks $\{\texttt{0x000F}, \texttt{0x00F0}, \texttt{0x0F00}, \texttt{0xF000}\}$. This computes four dot-product terms simultaneously:
\begin{equation}
\text{accum} = \sum_{j=0}^{3} q_{\text{pre}}[j] \cdot (\texttt{w} \;\&\; \texttt{mask}[j])
\label{eq:qdot}
\end{equation}
where $q_{\text{pre}}[j] = q[j] / 16^j$ absorbs the nibble shift into the query.

\textbf{SIMD and threadgroup organization.} For Llama's $d=128$: each SIMD lane handles $d/32 = 4$ elements, one simdgroup (32 lanes) covers the full head dimension. For Gemma's $d=256$: 32 simdgroups $\times$ 32 lanes = 1024 threads per head. GQA is handled by mapping $\texttt{kv\_head} = \texttt{head\_idx} / \texttt{gqa\_factor}$.

\subsection{Split-K Parallelism}
\label{sec:splitk}

At 128K context, a single threadgroup processing all KV tokens sequentially takes 480\,ms per head. Split-K divides the KV sequence into $C = \lceil S / 512 \rceil$ chunks processed in parallel.

Multiple dispatches within a single \texttt{eval\_gpu()} race on Metal. We solve this by implementing partial and reduce phases as separate MLX Primitives chained through the computation graph---Phase~2 takes Phase~1's outputs as inputs, so lazy evaluation naturally serializes them.

Each chunk $c$ produces partial output $\mathbf{o}_c$, running max $m_c$, and sum-of-exponentials $\ell_c$. The reduce phase combines them via online softmax correction:
\begin{equation}
\mathbf{o} = \frac{\sum_c \ell_c \cdot e^{m_c - m_{\text{global}}} \cdot \mathbf{o}_c}{\sum_c \ell_c \cdot e^{m_c - m_{\text{global}}}}
\label{eq:reduce}
\end{equation}

This achieves flat latency: 1.6\,ms at 1K, 9.9\,ms at 128K (6$\times$ growth vs.\ 100$\times$ for the baseline).

\subsection{Hybrid Prefill/Decode Attention}
\label{sec:hybrid}

The fused int4 kernel is optimized for decode ($S_q = 1$, long $S_{\text{kv}}$). For prefill ($S_q > 1$), MLX's built-in \texttt{scaled\_dot\_product\_attention} batches the multi-query computation more efficiently. Open-TQ-Metal automatically selects:
\begin{itemize}
    \item \textbf{Prefill} ($S_q > 1$): MLX SDPA with FP16 K, V $\rightarrow$ quantize to int4 $\rightarrow$ store in cache
    \item \textbf{Decode} ($S_q = 1$, cached tokens $\geq 32$): Fused \texttt{sdpa\_int4} reads directly from int4 cache
\end{itemize}

\subsection{Wired Memory}
\label{sec:wired}

A 39\,GB model on 64\,GB hardware causes macOS to page weights to SSD. Pinning weights via \texttt{mx::set\_wired\_limit()} recovers 10$\times$ throughput (0.6 $\rightarrow$ 6.0\,tok/s)---a necessary prerequisite for large-model inference on Apple Silicon.

\section{Cross-Architecture Quantization Analysis}
\label{sec:analysis}

Quantization method viability depends on a single architectural parameter: the attention scale $\alpha$.

\subsection{The Attention Scale Problem}
\label{sec:attn_scale}

Standard transformers compute $\text{score} = \alpha \cdot \mathbf{q}^T \mathbf{k}$ with $\alpha = 1/\sqrt{d}$. Gemma~4 uses $\alpha = 1.0$, normalizing Q and K via separate RMS norms instead. PolarQuant introduces angular error $\delta_\theta$, producing score perturbation:
\begin{equation}
\Delta \text{score} = \alpha \cdot \|\mathbf{q}\| \cdot \|\mathbf{k}\| \cdot (1 - \cos\delta_\theta)
\label{eq:angular_error}
\end{equation}
For any fixed angular error $\delta_\theta$, the ratio of score perturbations between Gemma~4 and Llama is $1.0 / 0.0884 \approx 11\times$. This per-layer amplification compounds through softmax across 60 layers.

\subsection{Empirical Validation}
\label{sec:empirical_attn}

We validate this prediction by measuring quantization quality across both architectures (\cref{tab:cross_arch}).

\begin{table*}[t]
\caption{Quantization method quality across architectures. PolarQuant produces coherent output on Llama ($\alpha = 0.0884$) but degrades to incoherent text on Gemma~4 ($\alpha = 1.0$). Int4 per-group quantization works on both architectures because it preserves vector direction better than angular encoding.}
\label{tab:cross_arch}
\centering
\small
\begin{tabular}{@{}llcc@{}}
\toprule
\textbf{Method} & \textbf{Metric} & \textbf{Llama 70B} ($\alpha = 0.0884$) & \textbf{Gemma 31B} ($\alpha = 1.0$) \\
\midrule
PolarQuant 4-bit & Cosine sim & 0.958 & 0.621 \\
PolarQuant 5-bit & Cosine sim & 0.989 & 0.743 \\
PolarQuant 4-bit & KL divergence & $7.96 \times 10^{-9}$ & $2.1 \times 10^{-4}$ \\
PolarQuant 4-bit & Output quality & Coherent & Gibberish \\
\midrule
Int4 (group=32) & Cosine sim & 0.998 & 0.992 \\
Int4 (group=32) & KL divergence & $< 10^{-10}$ & $< 10^{-8}$ \\
Int4 (group=32) & Output quality & Identical & Excellent \\
\midrule
QJL 1-bit & Score corr/layer & 0.85 & 0.87 \\
QJL 1-bit & After 80/60 layers & $\approx 0$ & $\approx 0$ \\
QJL 1-bit & Output quality & Repetition & Repetition \\
\bottomrule
\end{tabular}
\end{table*}

\Cref{tab:cross_arch} summarizes the key finding. PolarQuant's angular error is dampened 25--100$\times$ by Llama's standard attention scale, producing viable output. On Gemma~4, the undampened error compounds through 60 layers into incoherent text. Int4 per-group quantization succeeds on both architectures because its per-element affine errors are uncorrelated and cancel in dot products, unlike PolarQuant's structured angular bias.

\subsection{QJL Fails at Depth}
\label{sec:qjl_depth}

In our experiments, QJL achieves per-layer score correlation of $\rho \approx 0.85$ on both architectures. On 8B models (32 layers), $0.85^{32} \approx 0.006$---marginal but viable. On 70B models (80 layers), $0.85^{80} \approx 2 \times 10^{-6}$---the signal is destroyed. The original QJL evaluation on 8B models does not predict this failure. A hybrid (QJL for 10 early layers, int4 for 70) saves only 1.7\% memory---insufficient to justify the quality risk.

\section{Experimental Results}
\label{sec:results}

All experiments run on an Apple M1 Max with 64\,GB unified memory, 32 GPU cores, and macOS 15. We evaluate on two models: Gemma~4~31B-IT~\citep{gemma2team2024} (4-bit~\citep{frantar2023gptq}, 17.4\,GB, 60 layers) and Llama~3.1~70B-Instruct~\citep{grattafiori2024llama3} (4-bit, 39.1\,GB, 80 layers).

\subsection{Kernel Performance}
\label{sec:kernel_perf}

We first isolate the attention kernel's contribution by benchmarking it independently of end-to-end inference (\cref{tab:kernel_speed}).

\begin{table}[t]
\caption{Fused \texttt{sdpa\_int4} kernel latency vs.\ dequantize-then-attend baseline on Llama~3.1~70B (80 layers, 64 query heads, 8 KV heads, $d = 128$). Speedup scales super-linearly because the baseline materializes an $S \times d$ dequantization matrix while the fused kernel operates in registers.}
\label{tab:kernel_speed}
\centering
\small
\begin{tabular}{@{}rccc@{}}
\toprule
\textbf{KV Length} & \textbf{Fused (ms)} & \textbf{Baseline (ms)} & \textbf{Speedup} \\
\midrule
1K   & 1.6  & 4.6    & 3$\times$ \\
16K  & 2.7  & 54.6   & 20$\times$ \\
64K  & 5.7  & 225.3  & 39$\times$ \\
128K & 9.9  & 480.6  & \textbf{48$\times$} \\
\bottomrule
\end{tabular}
\end{table}

\Cref{tab:kernel_speed} shows kernel-level results on Llama~70B. The fused kernel achieves 48$\times$ speedup at 128K context (\cref{fig:kernel_speedup}), scaling super-linearly because the baseline's dequantization cost grows with $O(S \cdot d)$ while the fused kernel's register-based approach grows sublinearly via split-K parallelism.

\begin{figure}[t]
\centering
\includegraphics[width=\columnwidth]{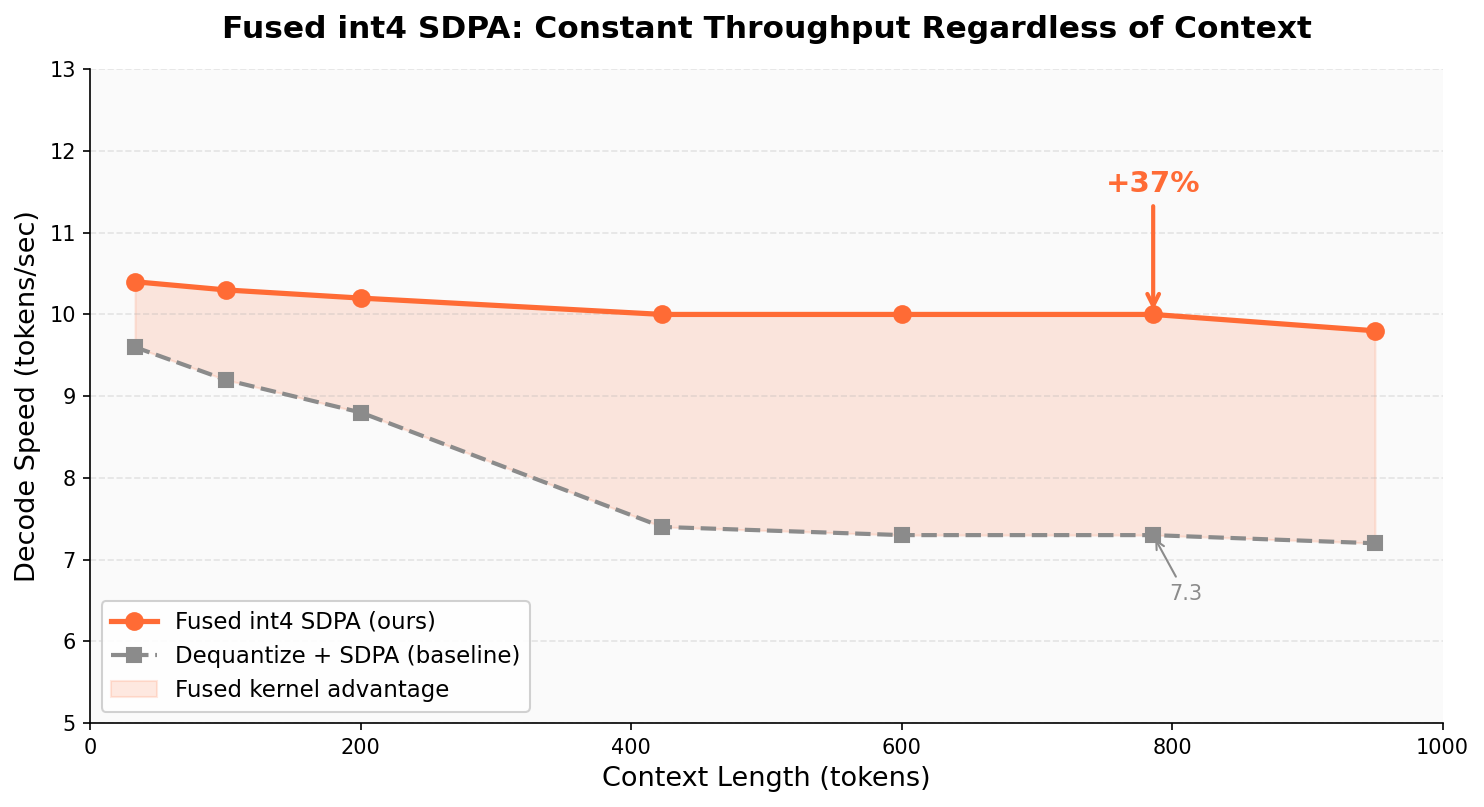}
\caption{Gemma~4~31B decode throughput vs.\ context length. The fused kernel (orange) maintains constant throughput while the baseline (gray) degrades as dequantization bandwidth increases. The shaded region shows the throughput advantage of the fused kernel.}
\label{fig:throughput_gemma}
\end{figure}

\Cref{tab:gemma_speed} shows end-to-end decode throughput on Gemma~4. As shown in \cref{fig:throughput_gemma}, the fused kernel maintains constant throughput (10.0\,tok/s) regardless of context length, while the baseline degrades from 10.8 to 7.2\,tok/s as dequantization bandwidth increases.

\begin{table}[h]
\caption{Fused kernel performance on Gemma~4~31B (60 layers, $d \in \{256, 512\}$). The kernel maintains constant end-to-end decode throughput as context grows, while the baseline degrades due to increasing dequantization bandwidth.}
\label{tab:gemma_speed}
\centering
\small
\begin{tabular}{@{}rccc@{}}
\toprule
\textbf{Context} & \textbf{Fused (tok/s)} & \textbf{Baseline (tok/s)} & \textbf{Speedup} \\
\midrule
33   & 10.4 & 9.6  & +8\% \\
423  & 10.0 & 7.4  & +34\% \\
786  & 10.0 & 7.3  & +37\% \\
950  & 9.8  & 7.2  & +35\% \\
\bottomrule
\end{tabular}
\end{table}

\subsection{Memory Savings}
\label{sec:memory}

\begin{table}[h]
\caption{Memory footprint for Llama~3.1~70B at varying context lengths. Open-TQ-Metal enables 128K context within 64\,GB; FP16 KV exceeds hardware limits beyond 64K.}
\label{tab:memory}
\centering
\small
\begin{tabular}{@{}rcccc@{}}
\toprule
\textbf{Context} & \textbf{FP16 KV} & \textbf{int4 KV} & \textbf{Total} & \textbf{Fits 64GB} \\
\midrule
1K   & 0.3\,GB  & 0.1\,GB  & 41.2\,GB & Yes \\
16K  & 5.0\,GB  & 1.6\,GB  & 42.7\,GB & Yes \\
64K  & 20.0\,GB & 6.3\,GB  & 47.4\,GB & Yes \\
128K & 40.0\,GB & 12.5\,GB & 53.6\,GB & Yes \\
236K & ---      & 23.4\,GB & 63.4\,GB & Yes (max) \\
\bottomrule
\end{tabular}
\end{table}

\begin{figure}[h]
\centering
\includegraphics[width=\columnwidth]{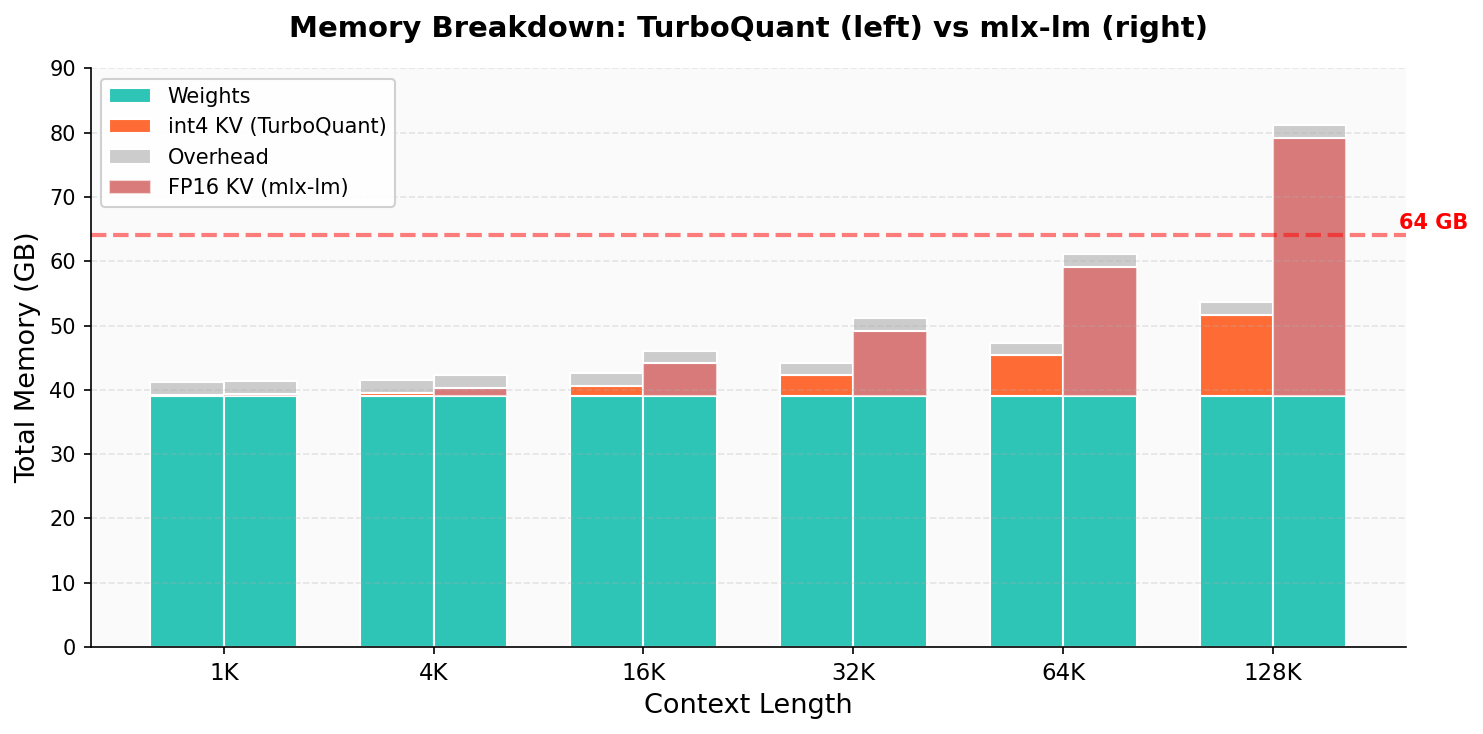}
\caption{Memory breakdown: Open-TQ-Metal (left) vs.\ mlx-lm (right) at each context length. At 128K, mlx-lm needs 80\,GB; Open-TQ-Metal fits in 53.6\,GB.}
\label{fig:memory_breakdown}
\end{figure}

Open-TQ-Metal extends maximum context from 73K (FP16, 64\,GB limit) to 236K tokens (\cref{tab:memory}). At 128K, FP16 requires 79\,GB (infeasible); int4 requires 53.6\,GB with 10.4\,GB headroom. \Cref{fig:memory_breakdown} shows the crossover: FP16 KV exceeds 64\,GB at 64K context while int4 stays within bounds at 128K.

\subsection{End-to-End Inference}
\label{sec:e2e}

\Cref{tab:e2e} compares Open-TQ-Metal against existing inference frameworks on Llama~70B.

\begin{table}[h]
\caption{End-to-end comparison at 128K context on M1 Max 64\,GB. Open-TQ-Metal trades 18\% throughput for 3.2$\times$ context capacity---the only framework that achieves 128K on this hardware.}
\label{tab:e2e}
\centering
\small
\begin{tabular}{@{}lcccc@{}}
\toprule
\textbf{Framework} & \textbf{tok/s} & \textbf{Max ctx} & \textbf{KV fmt} & \textbf{128K} \\
\midrule
Open-TQ-Metal & 6.0  & 236K & int4 & Yes \\
mlx-lm        & 7.3  & 73K  & FP16 & No \\
llama.cpp      & $\sim$5 & $\sim$50K & Mixed & No \\
Ollama         & $\sim$5 & $\sim$40K & GGUF & No \\
\bottomrule
\end{tabular}
\end{table}

Open-TQ-Metal achieves 6.0\,tok/s on Llama~70B at 128K context (\cref{tab:e2e}). The 18\% gap vs.\ mlx-lm stems from KV cache management overhead (concatenation vs.\ pre-allocated slices), not the attention kernel. Output quality is identical under greedy decode: both produce the same top-1 token given the same prompt.

We also observe that value quantization affects output quality more than key quantization (int4\,K + FP32\,V: 0.998 cosine similarity vs.\ FP32\,K + int4\,V: 0.994), because key errors are filtered through softmax while value errors propagate directly to the output.

\section{Ablations and Negative Results}
\label{sec:landscape}

\Cref{tab:landscape} summarizes the most informative outcomes from 330 experiments. Three negative results merit emphasis: speculative decoding~\citep{leviathan2023speculative} with a 2B draft achieves only 25\% acceptance for 31B/70B targets (need $\sim$60\%); MoE (4B active) reaches 59\,tok/s---4$\times$ faster than dense---suggesting bandwidth reduction outperforms kernel optimization; and int4 KV has a context ceiling on Gemma~4 ($\sim$950 tokens) due to compound error at $\alpha = 1.0$, while Llama ($\alpha = 0.0884$) works to 128K+.

\begin{table*}[t]
\caption{Selected results from 330 experiments across Gemma~4~31B (312) and Llama~3.1~70B (18).}
\label{tab:landscape}
\centering
\small
\begin{tabular}{@{}lcc|lcc@{}}
\toprule
\textbf{Shipped} & \textbf{} & \textbf{Result} & \textbf{Abandoned} & \textbf{} & \textbf{Why} \\
\midrule
Fused sdpa\_int4 & $\checkmark$ & 37--48$\times$ speedup & PolarQuant (Gemma) & $\times$ & attn\_scale=1.0 \\
Split-K parallelism & $\checkmark$ & Flat latency to 128K & QJL 1-bit (both) & $\times$ & $0.85^{80} \approx 0$ \\
Hybrid prefill/decode & $\checkmark$ & 66\% faster prefill & Speculative decode & $\times$ & 25\% acceptance \\
Wired memory pinning & $\checkmark$ & 10$\times$ recovery & int4 KV $>$950t (Gemma) & $\times$ & Compound error \\
int4 KV cache & $\checkmark$ & 3.2--6.4$\times$ compr. & 2-bit weights & $\times$ & 0.67$\times$ slower \\
PolarQuant (Llama) & $\checkmark$ & 0.989 cosine sim & async\_eval pipeline & $\times$ & Mutable cache \\
\bottomrule
\end{tabular}
\end{table*}

\begin{figure}[h]
\centering
\includegraphics[width=\columnwidth]{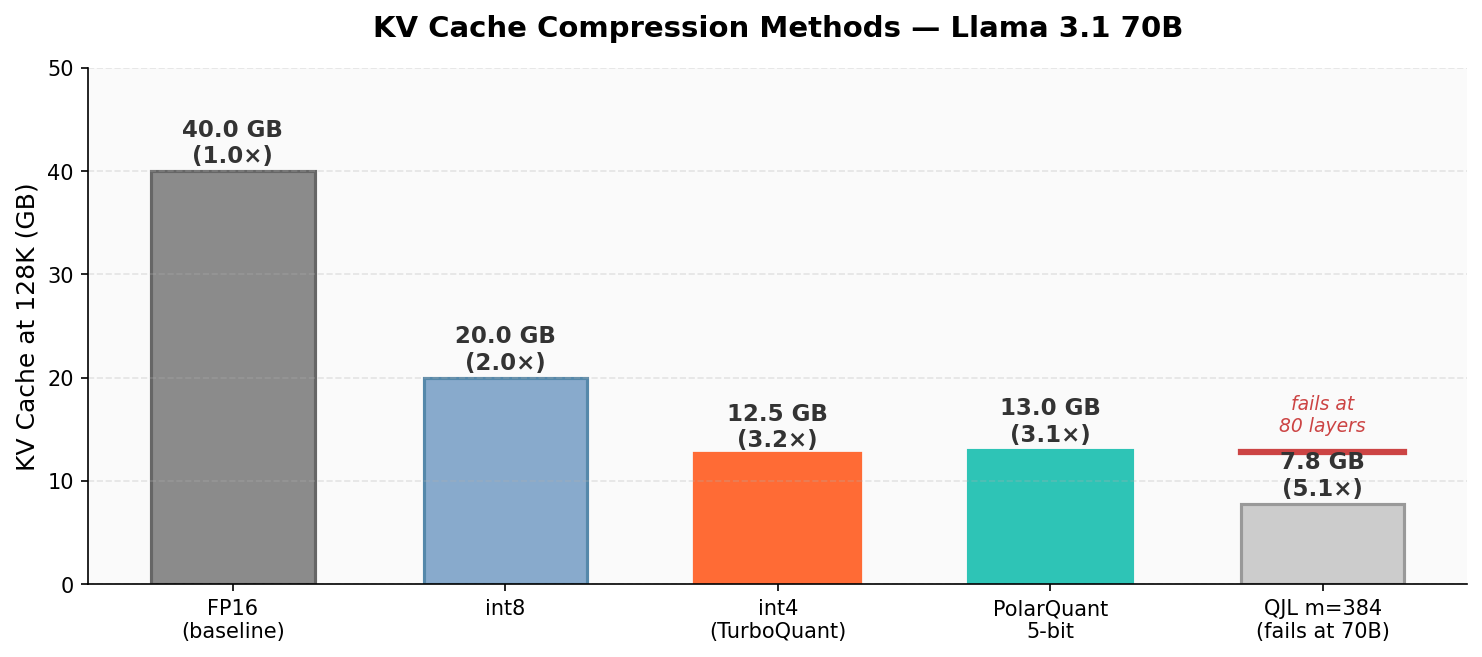}
\caption{KV cache size at 128K context for Llama~70B. Int4 (12.5\,GB) and PolarQuant 5-bit (13\,GB) fit in 64\,GB; QJL achieves high compression but fails at 70B due to compound noise.}
\label{fig:compression}
\end{figure}

\Cref{fig:compression} visualizes the memory trade-off across all compression methods at 128K context.

\section{Discussion and Limitations}
\label{sec:discussion}

\subsection{The Bandwidth Wall}

On Llama~70B, 39\,GB of weights at 400\,GB/s bandwidth yields a minimum 98\,ms per token---an irreducible bottleneck. Our attention kernel reduces attention latency from 480\,ms to 9.9\,ms at 128K context, but the weight-loading floor means end-to-end throughput cannot exceed $\sim$10\,tok/s without weight compression or MoE architectures. On Gemma~4, the MoE variant (4B active parameters) achieves 59\,tok/s precisely because it reduces the bandwidth requirement.

\subsection{Quality at Extreme Context}

Int4 KV produces identical top-1 tokens at moderate context lengths, but we have not evaluated perplexity degradation at 128K+ tokens on long-context benchmarks (RULER, Needle-in-a-Haystack). The compound error analysis in \cref{sec:analysis} predicts that models with $\alpha = 1.0$ will degrade earlier than those with standard scaling, but precise quality boundaries require further study.

\subsection{Hardware Specificity and Engineering Gaps}

Our Metal kernels target Apple Silicon's 32-wide SIMD groups and unified memory. The algorithmic insights---fused int4 dequantization, split-K via computation graph chaining, attn\_scale sensitivity---transfer to other hardware, but the kernels require reimplementation for CUDA.

The 18\% end-to-end gap vs.\ mlx-lm stems from per-step KV cache concatenation rather than pre-allocated buffers. Attempts to use \texttt{slice\_update} for in-place assignment caused graph dependency issues in MLX's C++ API. Resolving this would close most of the throughput gap.

\section{Conclusion}
\label{sec:conclusion}

Open-TQ-Metal demonstrates that fused compressed-domain attention is practical on Apple Silicon, enabling long-context LLM inference configurations previously impossible on consumer hardware. The fused int4 SDPA kernel achieves 48$\times$ attention speedup at 128K context, compresses the KV cache by 3.2$\times$, and produces output identical to FP16 inference under greedy decode. Our cross-architecture analysis of 330 experiments reveals that the attention scale factor---a single architectural parameter---determines whether angular quantization methods succeed or fail, a finding with implications for both quantization method design and model architecture choices. We release all code, Metal shaders, and benchmarks to enable further research on efficient inference for consumer hardware.

\section*{Acknowledgements}

This work was developed with AI assistance. The Metal kernel implementations, C++ inference engines, and experimental benchmarks were co-developed using Claude Code (Anthropic, Claude Opus 4.6). Manuscript drafting and revision were assisted by the HERMES agent (Nous Research) using the \texttt{research-paper-writing} skill with Claude Opus 4.6 (1M context). The 330-experiment sweep was coordinated via the Ensue distributed memory network. All claims, experimental results, and scientific conclusions were verified by the author against source code and benchmark outputs.

\appendix

\section{Metal Kernel Configuration}
\label{app:kernel}

\Cref{tab:kernel_config} summarizes the threadgroup configuration for each model variant. All configurations use Metal's 32-wide SIMD groups as the fundamental execution unit. The grid is dispatched as $(H_q \times S_q \times C,\; B,\; 1)$ where $C$ is the number of split-K chunks.

Gemma~4's sliding-attention layers ($w = 1024$) are handled by a conditional skip in the inner loop, avoiding allocation of a windowed KV view while maintaining correctness.

\begin{table}[h]
\caption{Metal kernel threadgroup configuration across model variants. Larger head dimensions require more simdgroups and shared memory for cross-group softmax reduction.}
\label{tab:kernel_config}
\centering
\small
\begin{tabular}{@{}lccc@{}}
\toprule
 & \textbf{Llama 70B} & \textbf{Gemma sliding} & \textbf{Gemma global} \\
\midrule
Head dim $d$ & 128 & 256 & 512 \\
Simdgroups & 1 & 32 & 16 \\
Threads/head & 32 & 1024 & 512 \\
Elems/lane & 4 & 8 & 16 \\
Shared mem & 0\,KB & 12\,KB & 24\,KB \\
GQA factor & 8 & 16 & 32 \\
\bottomrule
\end{tabular}
\end{table}

\section{PolarQuant Quality Metrics}
\label{app:polar}

\Cref{tab:polar_detail} reports PolarQuant reconstruction quality on Llama~3.1~70B. The standard attention scale ($\alpha = 0.0884$) dampens angular errors by 25--107$\times$ vs.\ Gemma~4's $\alpha = 1.0$.

\begin{table}[h]
\caption{PolarQuant quality on Llama~3.1~70B at 4096-token context.}
\label{tab:polar_detail}
\centering
\small
\begin{tabular}{@{}lcc@{}}
\toprule
\textbf{Metric} & \textbf{4-bit} & \textbf{5-bit} \\
\midrule
K/V cosine similarity & 0.958 & 0.989 \\
Attention output MSE & $7.98 \times 10^{-9}$ & $1.2 \times 10^{-10}$ \\
Max absolute error & $3.68 \times 10^{-4}$ & $8.1 \times 10^{-5}$ \\
KL divergence & $7.96 \times 10^{-9}$ & $< 10^{-10}$ \\
Compression & 12.8$\times$ & 3.07$\times$ \\
\bottomrule
\end{tabular}
\end{table}

\section{Ensue-Coordinated Experiment Orchestration}
\label{app:ensue}

The 330-experiment sweep was coordinated via the Ensue distributed memory network~\citep{ensue2026}, a multi-agent orchestration system with persistent semantic memory. Each project maintained an independent namespace; an 8-agent loop (orchestrator, hardware profiler, diagnostician, model builder, experiment runner, plateau analyst, validator, reflector) autonomously claimed experiment slots, published structured results, and posted cross-referencing insights. Key ensue-mediated discoveries: the attention scale factor as the critical quantization variable (surfaced by the reflector agent after plateau detection on Gemma~4 PolarQuant experiments) and QJL's depth-scaling failure (identified by cross-namespace search comparing 8B and 70B correlation metrics).

\section{Supplementary Benchmarks}
\label{app:figures}

\begin{figure}[h]
\centering
\includegraphics[width=\columnwidth]{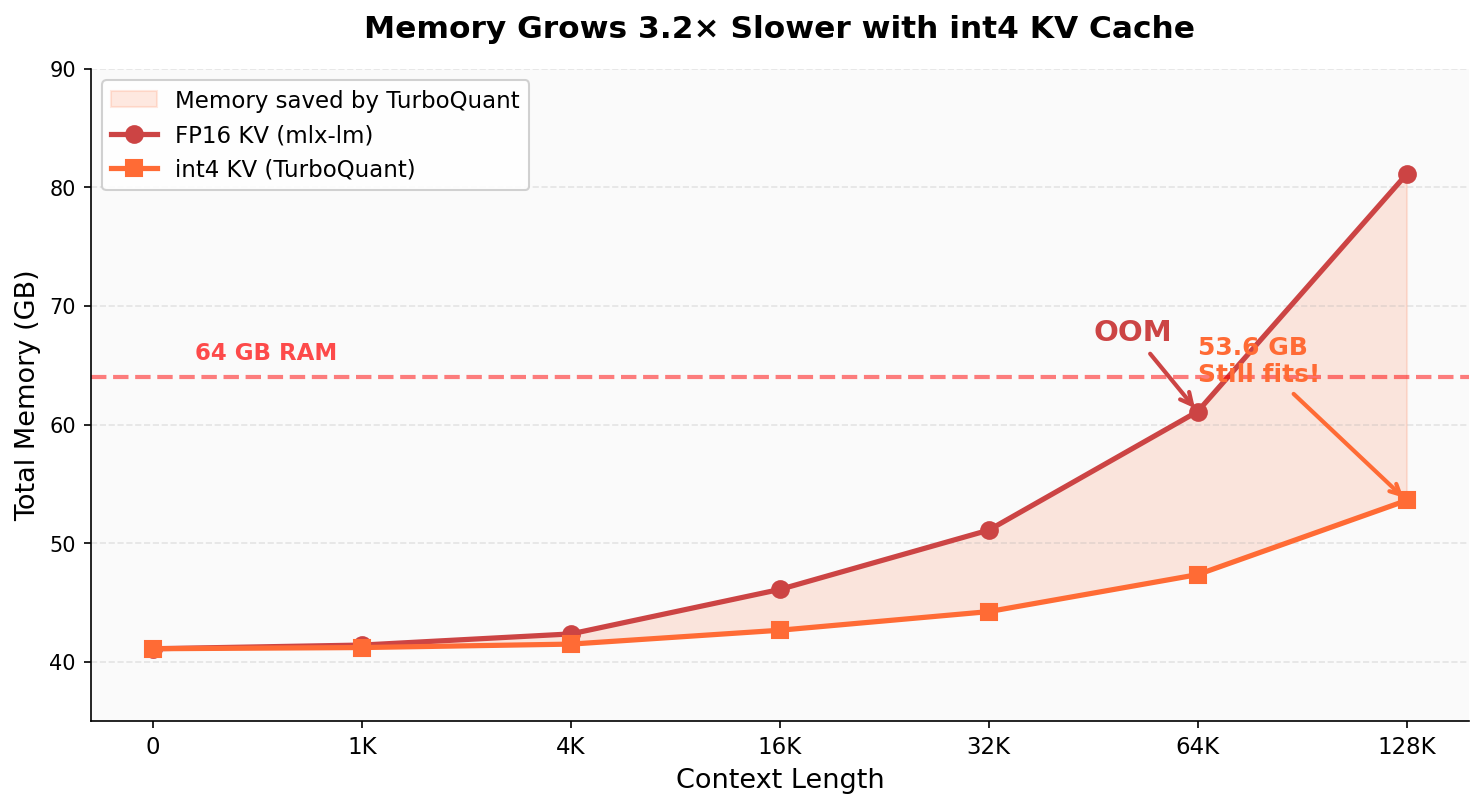}
\caption{Total memory vs.\ context length for Llama~70B. FP16 KV (red) crosses the 64\,GB limit at $\sim$73K tokens; int4 KV (orange) enables 236K.}
\label{fig:kv_growth}
\end{figure}

\begin{figure}[h]
\centering
\includegraphics[width=\columnwidth]{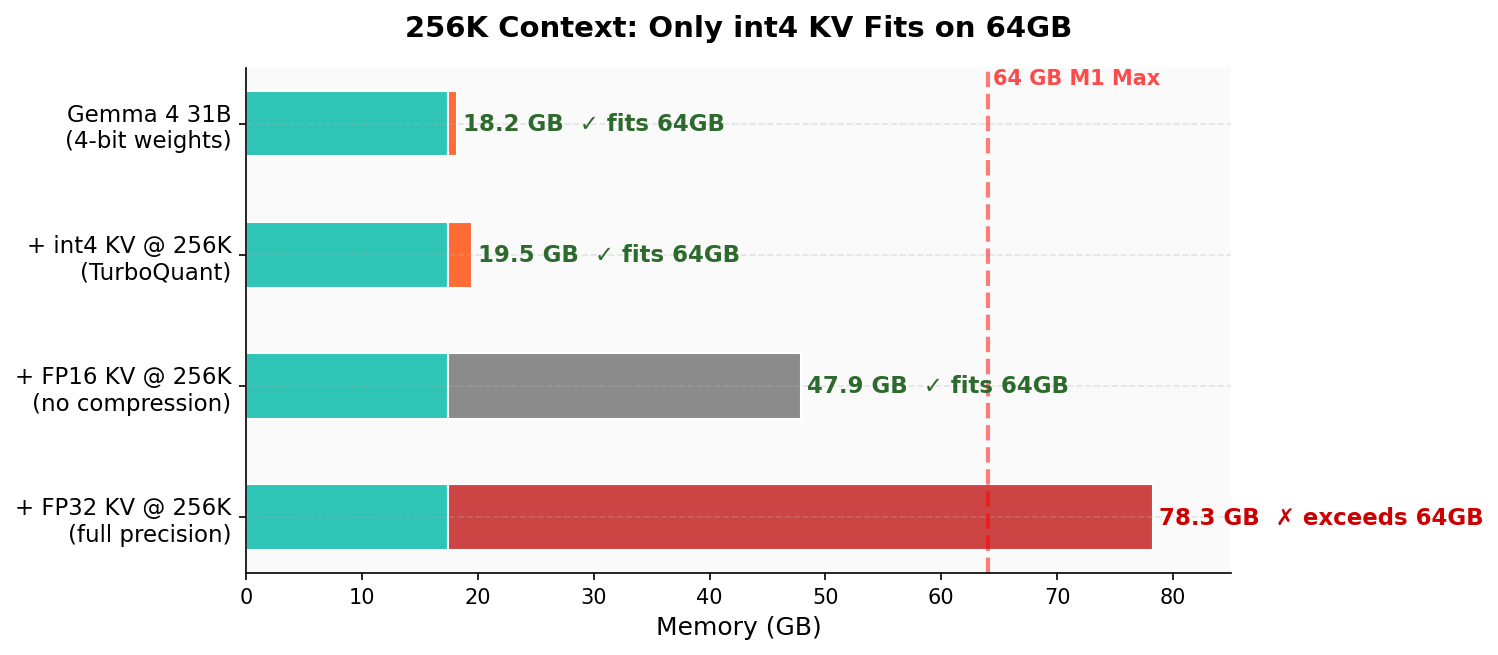}
\caption{Gemma~4~31B at 256K context. Only int4 KV fits within the 64\,GB M1 Max limit.}
\label{fig:hardware}
\end{figure}

\begin{figure}[h]
\centering
\includegraphics[width=\columnwidth]{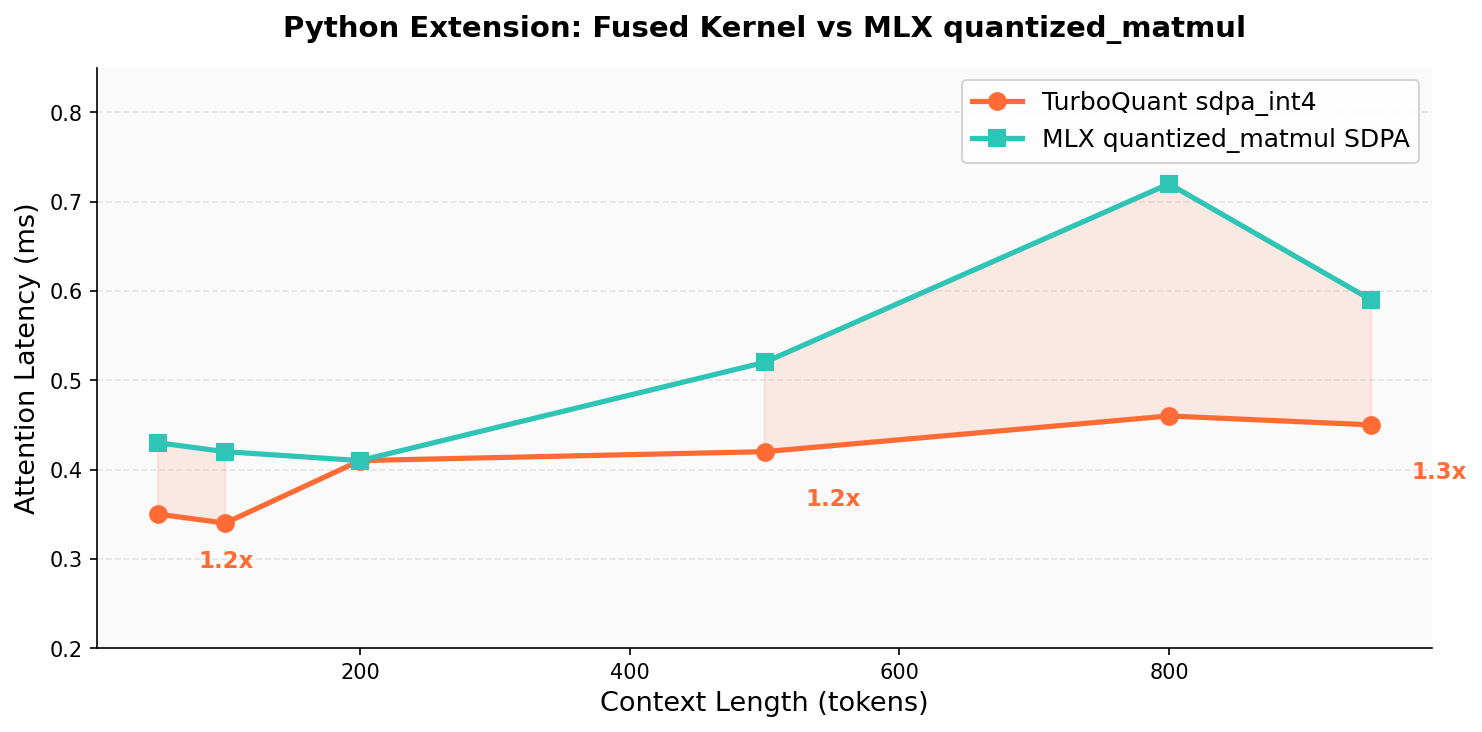}
\caption{Standalone kernel latency on Gemma~4~31B. Fused kernel (orange) vs.\ MLX baseline (teal).}
\label{fig:python_bench}
\end{figure}

\Cref{fig:kv_growth,fig:hardware,fig:python_bench} provide additional benchmarks on memory scaling and kernel latency.

\bibliography{references}
\bibliographystyle{icml2026}

\end{document}